\documentclass[10pt,twocolumn,letterpaper]{article}

\usepackage{iccv}

\usepackage{times}
\usepackage{epsfig}
\usepackage{graphicx}
\usepackage{amsmath}
\usepackage{amssymb}

\usepackage{xcolor}

\ificcvfinal\fi
\newcommand{\algname}{LERF}
\newcommand{\algfull}{Language Embedded Radiance Fields}
\newcommand{\todo}[1]{}
\newcommand{\revision}[1]{}
\renewcommand{\todo}[1]{{\color{red} TODO: {#1}}}
\renewcommand{\revision}[1]{{\color{orange} {#1}}}

\usepackage{subcaption}
\usepackage{xcolor}
\usepackage{booktabs}
\usepackage[font=small]{caption}

\usepackage{authblk}

\usepackage[breaklinks=true,bookmarks=false]{hyperref}

\iccvfinalcopy 

\def\iccvPaperID{9040} 

\ificcvfinal\fi

\begin{document}

\title{
LERF: Language Embedded Radiance Fields 
}

\author[]{Justin Kerr*}
\author[]{Chung Min Kim*}
\author[]{Ken Goldberg}
\author[]{Angjoo Kanazawa}
\author[]{Matthew Tancik}
\affil[]{UC Berkeley}


\twocolumn[{%
\renewcommand\twocolumn[1][]{#1}
\maketitle
\vspace{-2em}
\begin{center}
    \captionsetup{type=figure}
    \includegraphics[width=0.99\linewidth]{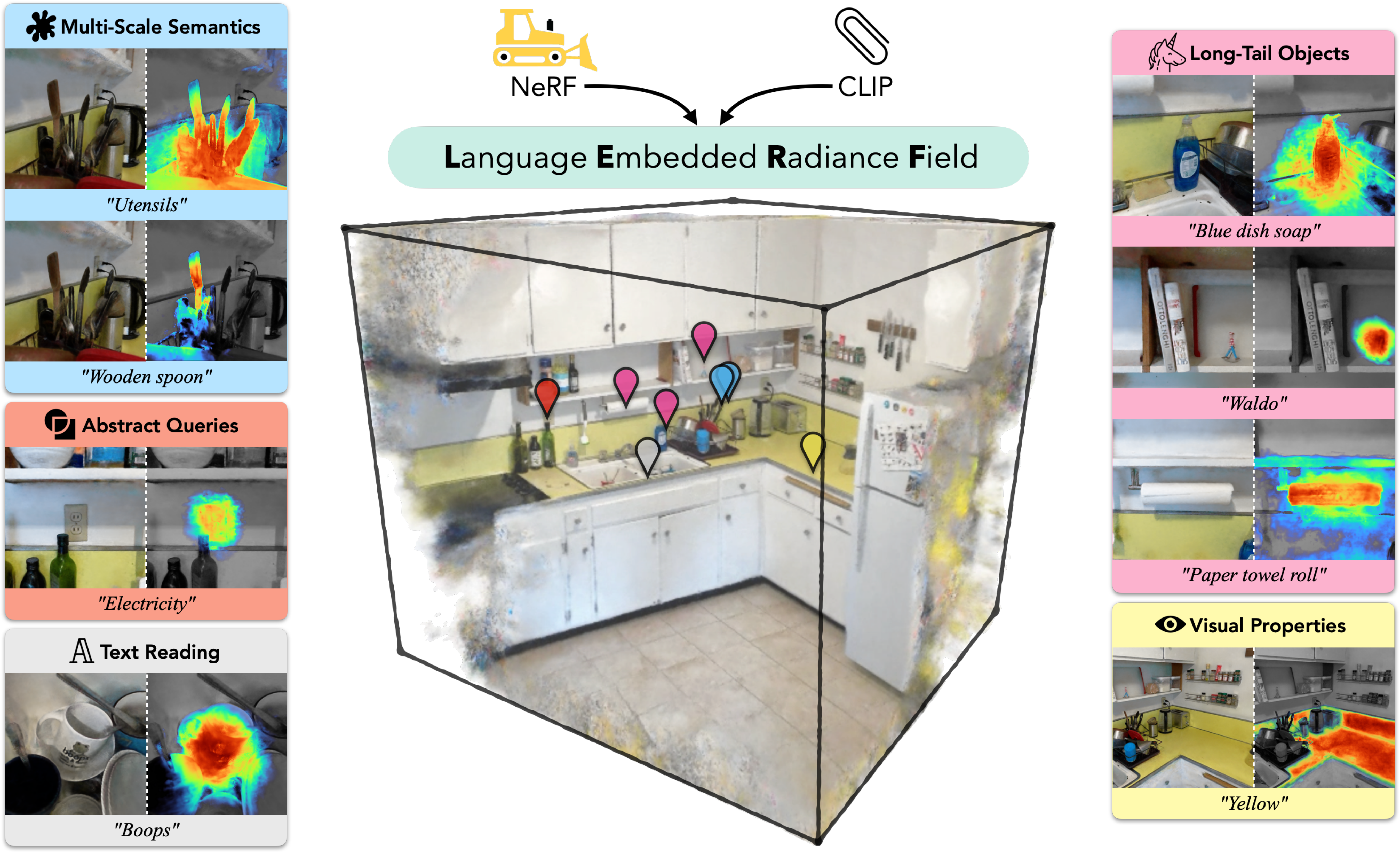}
    \captionof{figure}{\textbf{\algfull{} (\algname).} \algname{} grounds CLIP representations in a dense, multi-scale 3D field. 
    A \algname{} can be reconstructed from a hand-held phone capture within 45 minutes, then can render dense relevancy maps given textual queries interactively in real-time. 
    \algname{} enables a broad range of concepts to be queried via natural language, from abstract queries like \textit{``Electricity''}, visual properties like \textit{``Yellow''}, long-tail objects such as \textit{``Waldo''}, and even reading text like \textit{``Boops''} on the mug. 
        For each prompt, an RGB image and relevancy map are rendered focusing on the location with maximum relevancy activation.
    }
    \label{fig:splash_fig}
\end{center}

}]
\def\thefootnote{*}\footnotetext{Equal contribution, corresponding authors.}

\begin{abstract}
\vspace{-10pt}
Humans describe the physical world using natural language to refer to specific 3D locations based on a vast range of properties: visual appearance, semantics, abstract associations, or actionable affordances. 
In this work we propose \algfull{} (\algname{}s), a method for grounding language embeddings from off-the-shelf models like CLIP into NeRF, which enable these types of open-ended language queries in 3D.
 \algname{} learns a dense, multi-scale language field inside NeRF by 
 volume rendering CLIP embeddings along training rays, supervising these embeddings across training views to provide multi-view consistency and smooth the underlying language field. 
 After optimization, \algname{} can extract 3D relevancy maps for a broad range of language prompts interactively in real-time, which has potential use cases in robotics, understanding vision-language models, and interacting with 3D scenes. \algname{} enables pixel-aligned, zero-shot queries on the distilled 3D CLIP embeddings without relying on region proposals or masks, supporting long-tail open-vocabulary queries hierarchically across the volume. See the project website at: \url{https://lerf.io}.
\end{abstract}
\section{Introduction} 
\label{sec:introduction}

Neural Radiance Fields (NeRFs)~\cite{mildenhall2020nerf} have emerged as a powerful technique for capturing photorealistic digital representations of intricate real-world 3D scenes. 
However, the immediate output of NeRFs is nothing but a colorful density field, devoid of meaning or context, which inhibits building interfaces for interacting with the resulting 3D scenes. 

Natural language is an intuitive interface for interacting with a 3D scene. Consider the capture of a  kitchen in Figure~\ref{fig:splash_fig}. Imagine being able to navigate this kitchen by asking where the \textit{``utensils''} are, or more specifically for a tool that you could use for \textit{``stirring''}, and even for your favorite mug with a specific logo on it --- all through the comfort and familiarity of everyday conversation.
This requires not only the capacity to handle natural language input queries but also the ability to incorporate semantics at multiple scales and relate to long-tail and abstract concepts.

In this work, we propose Language Embedded Radiance Fields (\algname{}), a novel approach that grounds language within NeRF by optimizing embeddings from an off-the-shelf vision-language model like CLIP into 3D scenes. Notably, \algname{} utilizes CLIP directly without the need for finetuning through datasets like COCO or reliance on mask region proposals, which limits the ability to capture a wide range of semantics. Because \algname{} preserves the integrity of CLIP embeddings at multiple scales, it is able to handle a broad range of language queries, including visual properties (\textit{``yellow''}), abstract concepts (\textit{``electricity''}), text (\textit{``boops''}), and long-tail objects (\textit{``waldo''}) as illustrated in Figure~\ref{fig:splash_fig}.

We construct a \algname{} by optimizing a language field jointly with NeRF, which takes both position and physical scale as input and outputs a single CLIP vector. During training, the field is supervised using a multi-scale feature pyramid that contains CLIP embeddings generated from image crops of training views. This allows the CLIP encoder to capture different scales of image context, thus associating the same 3D location with distinct language embeddings at different scales (\eg \textit{``utensils"} \vs \textit{``wooden spoon"}). The language field can be queried at arbitrary scales during test time to obtain 3D relevancy maps. To regularize the optimized language field, self-supervised DINO~\cite{dino} features are  also incorporated through a shared bottleneck. 

\algname{} offers an added benefit: since we extract CLIP embeddings from multiple views over multiple scales, the relevancy maps of text queries obtained through our 3D CLIP embedding are more localized compared to those obtained via 2D CLIP embeddings. By definition, they are also 3D consistent, enabling queries directly in the 3D fields without having to render to multiple views.

\algname{} can be trained without significantly slowing down the base NeRF implementation. Upon completion of the training process, \algname{} allows for the generation of 3D relevancy maps for a wide range of language prompts in real-time. We evaluate the capabilities of \algname{} on a set of hand-held captured in-the-wild scenes and find it can localize both fine-grained queries relating to highly specific parts of geometry (\textit{``fingers"}), or abstract queries relating to multiple objects (\textit{``cartoon"}). \algname{} produces view-consistent relevancy maps in 3D across a wide range of queries and scenes, which are best viewed in videos on our website. We also provide quantitative evaluations against popular open-vocab detectors LSeg~\cite{li2022language_lseg} and OWL-ViT~\cite{minderer2022simple_owlvit}, by distilling LSeg features into 3D~\cite{kobayashi2022decomposing} and querying OWL-ViT from rendered novel views. Our results suggest that features in 3D from \algname{} can localize a wide variety of queries across in-the-wild scenes. The zero-shot capabilities of \algname{} leads to potential use cases in robotics, analyzing vision-language models, and interacting with 3D scenes. Code and data will be made available at \url{https://lerf.io}.
\begin{figure*}
    \centering
    \includegraphics[width=\linewidth]{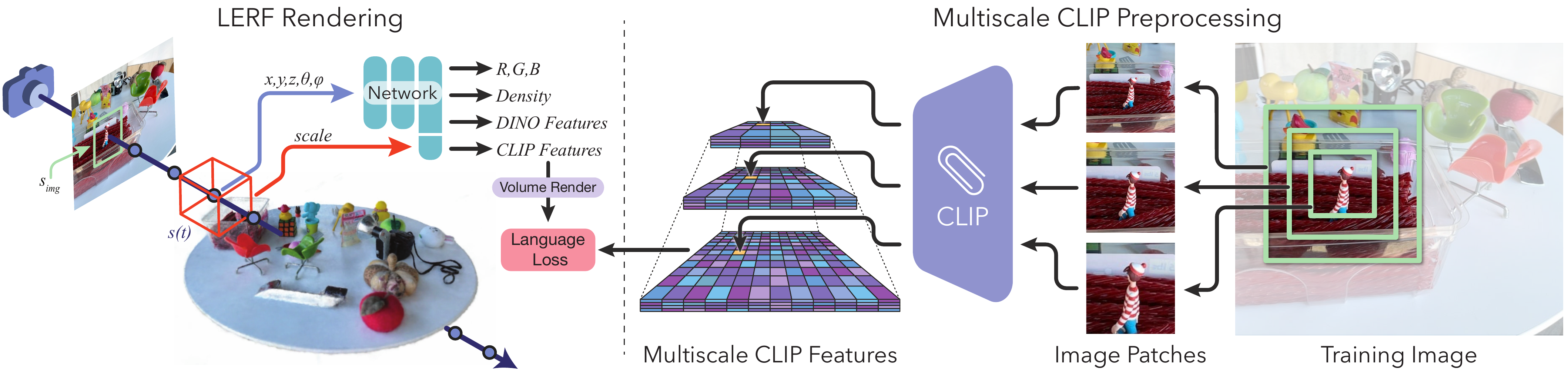}
    \caption{
    \textbf{\algname{} Optimization:} 
    \textit{Left}: \algname{} represents a field of 3D volumes, parameterized by position $x,y,z$ and scale $s$ (orange cube). To render a CLIP embedding along a ray, the field is sampled and averaged according to NeRF's volume rendering weights. Physical scale corresponds to an image scale $s_\text{img}$ via projective geometry. \textit{Right}: We pre-compute a multi-scale feature pyramid of CLIP embeddings over training views, and during training interpolate this pyramid with $s_\text{img}$ and the ray's pixel location to obtain CLIP supervision. The CLIP loss maximizes cosine similarity, and other outputs are supervised with mean squared-error using standard per-pixel rendering.}
    \vspace{-12pt}
    \label{fig:pipeline}
\end{figure*}
\section{Related Work}
\paragraph{Open-Vocabulary Object Detection}
A number of approaches study detecting objects in 2D images given natural language prompts. These lie on a spectrum from purely zero-shot to fully trained on segmentation datasets.
LSeg~\cite{li2022language_lseg} trains a 
2D image encoder on labeled segmentation datasets, which outputs pixel-wise embeddings that best match the CLIP text embedding of the segmentation label at that given pixel. CRIS~\cite{wang2022cris} and CLIPSeg~\cite{luddecke2022image_clipseg} train a 2D image decoder to output a relevancy map based on the query CLIP embedding and the intermediate outputs of the pretrained CLIP image encoder. However, such fine-tuning approaches tend to lose significant language capabilities by training on a smaller dataset.

Another common approach for 2D images is a two-stage framework wherein class-agnostic region or mask proposals direct where to query open-vocabulary classification models. OpenSeg~\cite{ghiasi2021open_openseg} simultaneously learns a mask prediction model while predicting the text embeddings for each mask, while ViLD~\cite{gu2021open_vild} directly uses CLIP~\cite{radford2021learning} to classify 2D regions from class-agnostic mask proposal networks. Detic~\cite{zhou2022detecting_detic} builds on existing two-stage object detector approaches, but demonstrates a greater generalization ability by allowing detector classifiers to train with image classification data. OWL-ViT~\cite{minderer2022simple_owlvit} attaches lightweight object classification and localization heads after a pre-trained 2D image encoder. Although region proposal-based approaches can leverage more detection data, these proposal generators still tend to output within the training set distribution. Consequently, as noted by the authors of OV-Seg~\cite{liang2022open_ovseg}, such networks often face difficulties in accurately segmenting unlabeled hierarchical components of the original masks, such as object parts. \algname{} strives to avoid region proposals by incorporating language embeddings in a dense, 3D, multi-scale field which allows hierarchical text queries.

Grad-CAM~\cite{selvaraju2017grad} and attention-based methods \cite{chefer2021generic} provide a relevancy mapping between 2D images and text in vision-language models (e.g., CLIP). Works such as Semantic Abstraction~\cite{ha2022semantic} have shown that these frameworks can be used to detect long-tail objects for scene understanding. Outputs from \algname{} are most similar in spirit to these methods, outputting a 3D relevancy score given a query. However, \algname{} builds a 3D representation that can be queried with different text prompts without reconstructing the underlying representation each time, and in addition fuses multiple viewpoints into a single shared scene representation, rather than operating per-image.
 
\paragraph{Distilling 2D Features into NeRF}
NeRF has an attractive property of averaging information across multiple views. Several prior works leverage this to improve the quality of 2D semantics, segmentations, or feature vectors by distilling them into 3D. Semantic NeRF~\cite{semantic_nerf} and Panoptic Lifting~\cite{panoptic_lifting} embed semantic information from semantic segmentation networks into 3D, showing that combining noisy or sparse labels in 3D can result in crisp 3D segmentations. This concept has been applied to segmenting objects with extremely sparse user input scribbles of foreground-background masks~\cite{ren-cvpr2022-nvos}. Our approach draws inspiration from these works by averaging multiple potentially noisy language embeddings over input views. 

Distilled Feature Fields~\cite{kobayashi2022decomposing} and Neural Feature Fusion Fields~\cite{tschernezki22neural} explore embedding pixel-aligned feature vectors from LSeg or DINO~\cite{dino} into a NeRF, and show they can be used for 3D manipulations of the underlying geometry. \algname{} similarly embeds feature vectors into NeRF, but also demonstrates a method to distill non pixel-aligned embeddings (e.g., from CLIP) into 3D without fine-tuning.
\begin{figure*}
    \centering
    {
    \includegraphics[width=\linewidth]{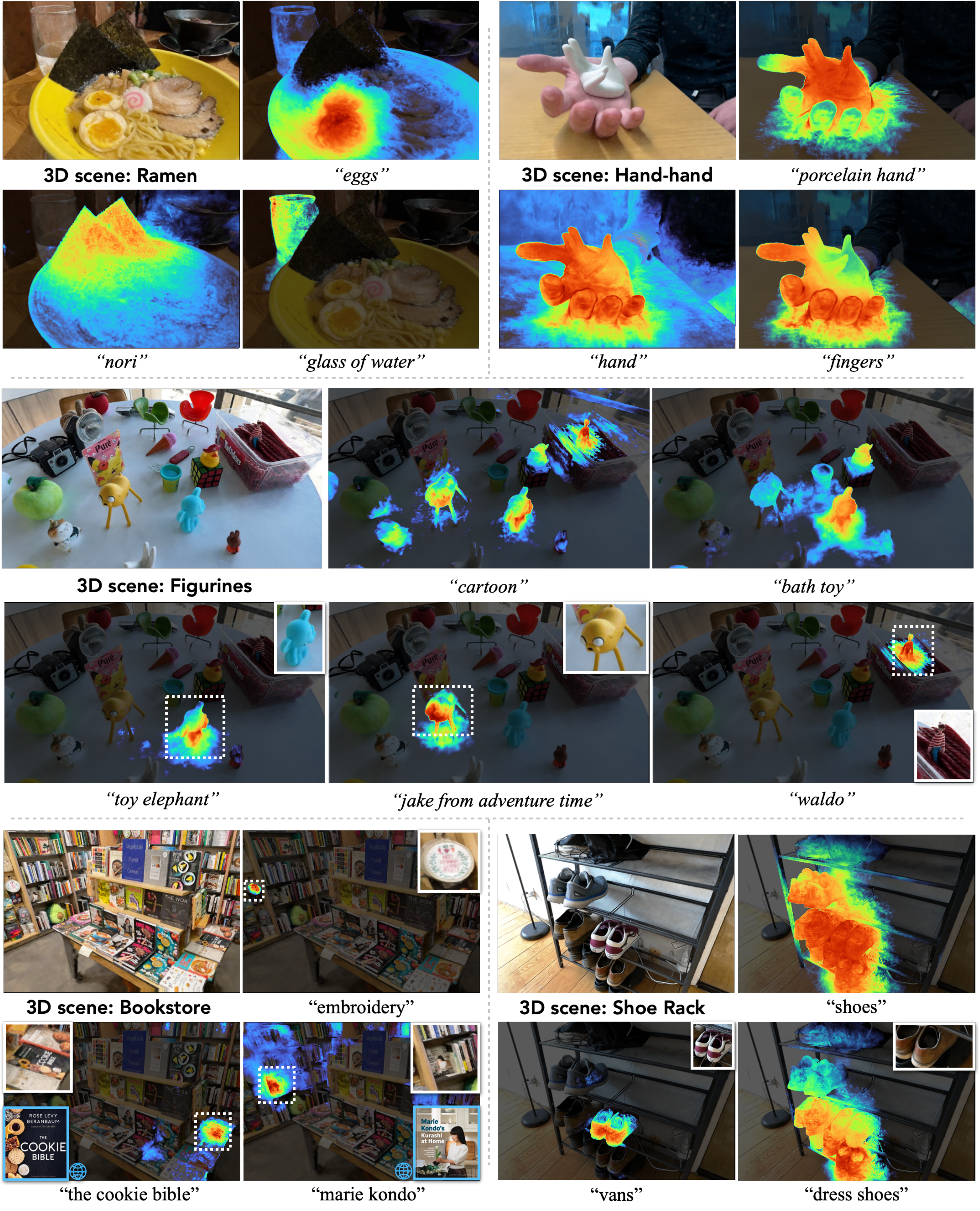}
    \caption{
    \textbf{Results with \algname{} for 5 in-the-wild scenes.}
    Each image shows a visual rendering of the \algname{} (Sec.~\ref{sec:methods}), along with relevancy renderings (Sec.~\ref{sec:querying}) for each text query and a cropped view of the activated region. For the bookstore scene, the original book cover images are shown in blue with a globe icon. See Sec.~\ref{sec:qualitative} for discussion and details on relevancy visualization.
    }
    \label{fig:results}
    }
\end{figure*}
\paragraph{3D Language Grounding}
Incorporating language into 3D has been explored in a wide range of contexts: 3D visual question answering~\cite{gordon2018iqa,azuma2022scanqa,cascante2022simvqa} leverage 3D information to extract answers to queries about the environment. In addition, incorporating language with shape information can improve object recognition via text~\cite{corona2022voxel,thomason2022language}. 

\algname{} is more similar to 3D scene representations in robotics which fuse vision-language embeddings to support natural language interaction. VL-Maps~\cite{vlmaps} and OpenScene~\cite{peng2022openscene} build a 3D volume of language features which can be queried for navigation tasks, by using pre-trained pixel-aligned language encoders~\cite{ghiasi2021open_openseg, li2022language_lseg}.
In \algname{}, we compare against one such encoder, LSeg, in 3D and find it loses significant expressive capability compared to CLIP.

CLIP-Fields~\cite{shafiullah2022clip} and NLMaps-SayCan~\cite{chen2022open} fuse CLIP embeddings of crops into pointclouds, using a contrastively supervised field and classical pointcloud fusion respectively. In CLIP-Fields, the crop locations are guided by Detic~\cite{zhou2022detecting_detic}. On the other hand, NLMaps-SayCan relies on region proposal networks. These maps are sparser than \algname{} as they primarily query CLIP on detected objects rather than densely throughout views of the scene. Concurrent work ConceptFusion~\cite{conceptfusion} fuses CLIP features more densely in RGBD pointclouds, using Mask2Former~\cite{cheng2022masked} to predict regions of interest, meaning it can lose objects which are out of distribution to Mask2Former's training set. In contrast, \algname{} does not use region or mask proposals.

\algname{} contributes a new dense, volumetric interface for 3D text queries which can integrate with a broad range of downstream applications of 3D language, improving the resolution and fidelity at which these methods can query the environment when multi-view inputs are available. This is enabled by the smoothing behavior of embedding potentially noisy feature vectors from multiple views into the dense \algname{} structure.

\section{\algfull{}}
\label{sec:methods}

Given a set of calibrated input images, we ground CLIP embeddings into a 3D field within NeRF. However, querying a CLIP embedding for a single 3D point is ambiguous, as CLIP is inherently a global image embedding and not conducive to pixel-aligned feature extraction. To account for this property, we propose a novel approach that involves learning a field of language embeddings over \textit{volumes} centered at the sample point. Specifically, the output of this field is the average CLIP embedding across all training views of image crops containing the specified volume. By reframing the query from points to volumes, we can effectively supervise a dense field from coarse crops of input images, which can be rendered in a pixel-aligned manner by conditioning on a given volume scale. 

\subsection{\algname{} Volumetric Rendering}
NeRF takes in a position $\vec x$ and view direction $\vec d$ and outputs color $\vec c$ and density $\sigma$. Samples of these values can be composited along a ray to produce a pixel's color. To create \algname{}, we augment NeRF's outputs with a language embedding $F_\text{lang}(\vec x,s)\in \mathbb{R}^d$, which takes an input position $\vec x$ and physical scale $s$, and outputs a $d$-dimensional language embedding. We choose this output to be view-\textit{independent}, since the semantic meaning of a location should be invariant to viewing angle. This allows multiple views to contribute to the same field input, averaging their embeddings together. The scale $s$ represents the side length in world coordinates of a cube centered at $\vec x$, and is analogous to how Mip-NeRF\cite{mipnerf,mipnerf360} incorporates different scales via integrated positional encodings.

Rendering color and density from \algname{} remains exactly the same as NeRF. To render language embeddings into an image, we adopt a similar technique as prior work\cite{kobayashi2022decomposing,semantic_nerf} to render language embeddings along a ray $\vec{r}(t)=\vec{o}_t+t\vec{d}$. However, since \algname{} is a field over \textit{volumes}, not points, we must also define a scale parameter for each position along the ray. We achieve this by fixing an initial scale in the image plane $s_\text{img}$ and define $s(t)$ to increase proportionally with focal length and sample distance from the ray origin: $s(t)=s_\text{img}*f_{xy}/t$ (Fig.~\ref{fig:pipeline}, left). Geometrically, this represents a frustrum along the ray. We calculate rendering weights as in NeRF: $T(t)=\int_t\text{exp}\left(-\sigma(s)ds\right) ,~w(t)=\int_tT(t)\sigma(t)dt$, then integrate the \algname{} to obtain raw language outputs: 
$\hat \phi_\text{lang}=\int_t w(t) F_\text{lang}\left(r(t),s(t)\right)dt$, and finally normalize each embedding to the unit sphere as in CLIP: $\phi_\text{lang}=\hat \phi_\text{lang}/||\hat \phi_\text{lang}||$.  We find that spherical interpolation between samples on a ray is unnecessary because non-zero weight samples along a ray tend to be spatially close, and instead opt for a weighted euclidean average followed by normalization for implementation simplicity. 

\begin{figure}
    \centering
    \includegraphics[width=0.95\linewidth]{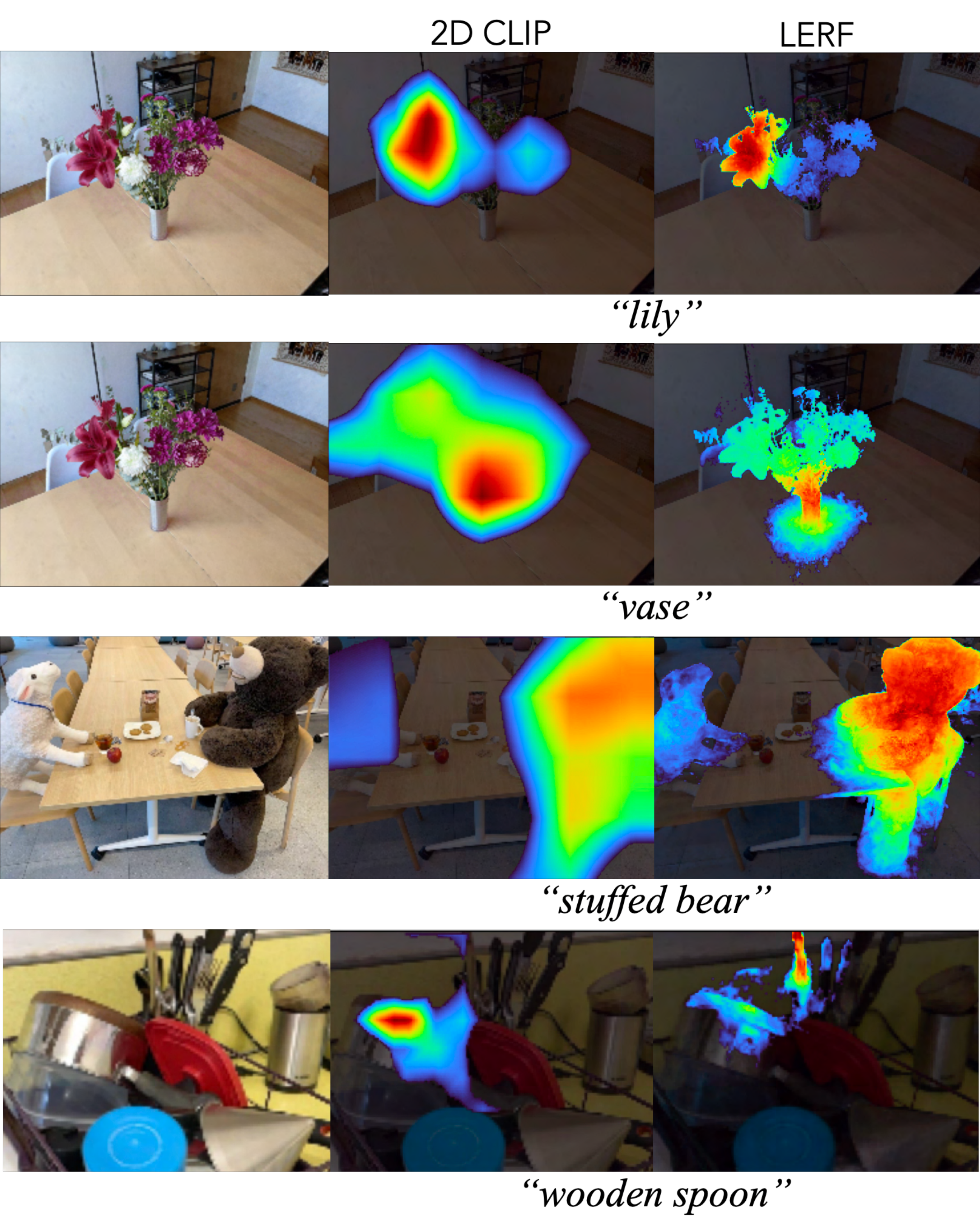}
    \caption{\textbf{2D CLIP vs \algname{}}: The left visualizes similarity interpolated over patchwise CLIP embeddings, and the right rendered from \algname{}. Because volumetric language rendering incorporates information from multiple views, 3D relevancy activation maps have better alignment with the underlying scene geometry.}
    \vspace{-15pt}
    \label{fig:clip-3d-comparison}
\end{figure}
\subsection{Multi-Scale Supervision}
To supervise language field outputs $F_\text{lang}$, recall that we can only query language embeddings over image patches, not pixels. Therefore, to supervise the multi-scale \algname{}, we supervise each rendered frustrum with an image crop of size $s_\text{img}$ centered at the image pixel where the ray originated. In practice, re-computing a CLIP embedding for each ray during \algname{} optimization would be prohibitively expensive, so instead we pre-compute an image pyramid over multiple image crop scales and store the CLIP embeddings of each crop (Fig~\ref{fig:pipeline}, right). This pyramid has $n$ layers sampled between $s_\text{min}$ and $s_\text{max}$, with each crop arranged in a grid with 50\% overlap between crops. 

During training, we randomly sample ray origins uniformly throughout input views, and uniformly randomly select $s_\text{img}\in(s_\text{min},s_\text{max})$ for each. Since these samples don't necessarily fall in the center of a crop in this image pyramid, 
we perform trilinear interpolation between the embeddings from the 4 nearest crops for the scale above and below
to produce the final ground truth embedding $\phi_\text{lang}^\text{gt}$. We minimize a loss between rendered and ground truth embeddings which maximizes cosine similarity between the two, scaling by a constant $\lambda_\text{lang}$ (Sec.~\ref{sec:dino}): $L_\text{lang}=-\lambda_\text{lang}\phi_\text{lang}\cdot \phi_\text{lang}^\text{gt}$.

\begin{figure}
    \centering
    \includegraphics[width=\linewidth]{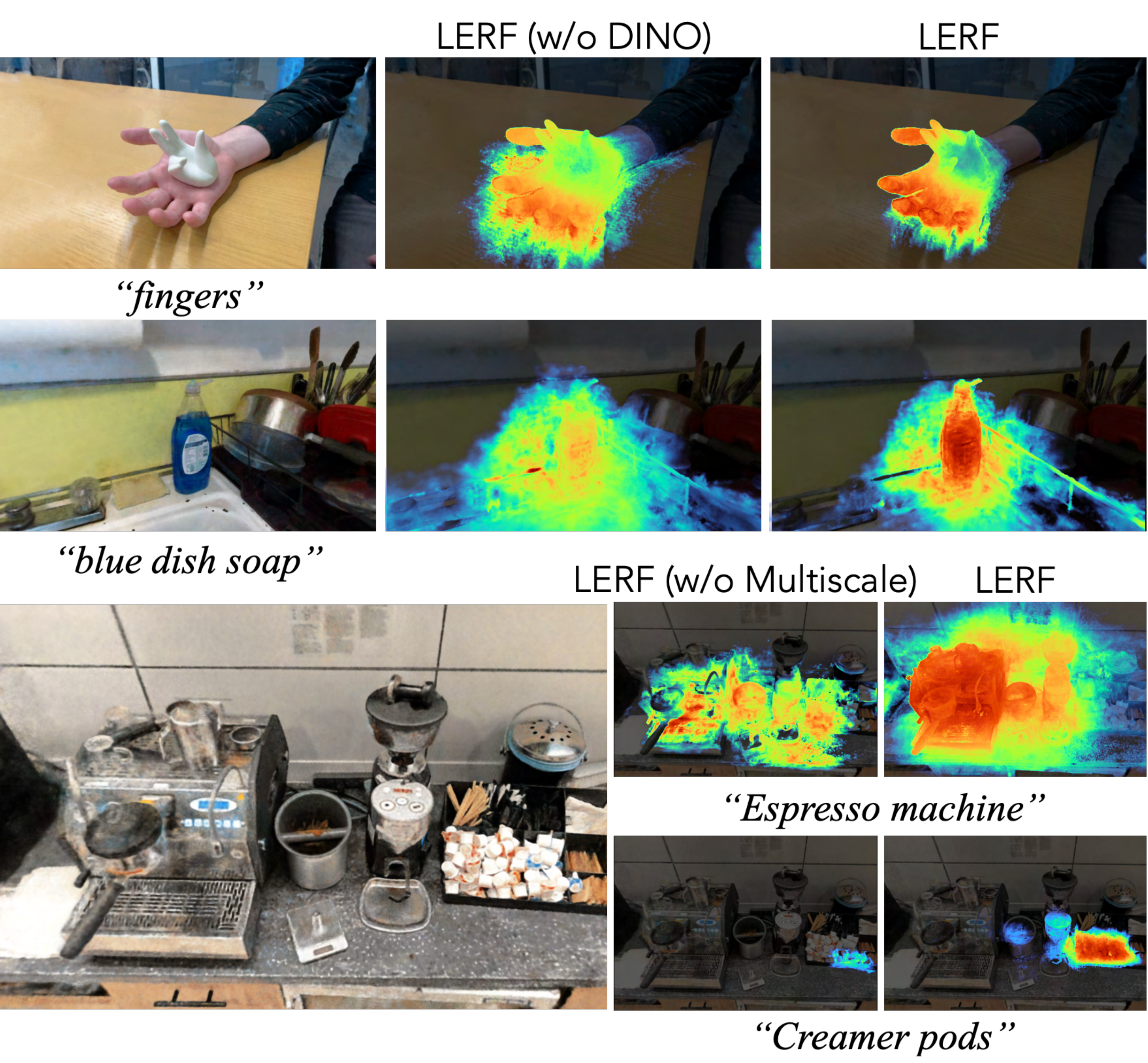}
    \caption{\textbf{Ablations}: We ablate DINO regularization and multi-scale training (Sec.~\ref{sec:ablations}), and highlight qualitative degradation in relevancy maps here.}
    \label{fig:ablations}
\end{figure}
\subsection{DINO Regularization} 
\label{sec:dino}
Na\"ively implementing \algname{} as described produces cohesive results, but the resulting relevancy maps can sometimes be patchy and contain outliers in regions with few views or little foreground-background separation (Fig. \ref{fig:ablations}). 

To mitigate this, we additionally train a field $F_{\text{dino}}(\vec x)$ which outputs a DINO\cite{dino} feature at each point. DINO has been shown to exhibit emergent object decomposition properties despite training on no labels\cite{amir2021deep}, and additionally distills well into 3D fields\cite{kobayashi2022decomposing}, making it a good candidate for grouping language in 3D  without relying on labeled data or imparting too strict a prior. Because DINO outputs pixel-aligned features, $F_\text{dino}$ does not take in scale as an input, and is directly supervised for each ray with the DINO feature it corresponds to. We render $\phi_\text{dino}$ identically to $\phi_\text{lang}$ except without normalizing to a unit sphere, and supervise it with MSE loss on ground-truth DINO features.
DINO is used explicitly during inference, and only serves as an extra regularizer during training since CLIP and DINO output heads share an architectural backbone.
\subsection{Field Architecture}
Intuitively, optimizing a language embedding in 3D should not influence the distribution of density in the underlying scene representation. We capture this inductive bias in \algname{} by training two separate networks: one for feature vectors (DINO, CLIP), and the other for standard NeRF outputs (color, density). Gradients from $L_\text{lang}$ and $L_\text{dino}$ \textit{do not affect} the NeRF outputs, and can be viewed as jointly optimizing a language field in conjunction with a radiance field.

We represent both fields with a multi-resolution hashgrid \cite{muller2022instant}. The language hashgrid has two output MLPs for CLIP and DINO respectively. Scale $s$ is passed into the CLIP MLP as an extra input in addition to the concatenated hashgrid features. We adopt the Nerfacto method from Nerfstudio~\cite{tancik2023nerfstudio} as the backbone for our approach, leveraging the same proposal sampling, scene contraction, and appearance embeddings

\subsection{Querying \algname{}}
\label{sec:querying}
Often, language models like CLIP are evaluated on zero-shot classification, where a category is selected from a list guaranteed to include the correct category~\cite{radford2021learning}. However, in practical usage of \algname{} on in-the-wild scenes, an exhaustive list of categories is not available. We view the open-endedness and ambiguity of natural language as a benefit, and propose a method to query 3D relevancy maps from the \algname{} given an arbitrary text query. Querying \algname{} has two parts: 1) obtaining a relevancy score for a rendered embedding and 2) automatically selecting a scale $s$ given the prompt.

\textbf{Relevancy Score}: To assign each rendered language embedding $\phi_\text{lang}$ a score, we compute the CLIP embedding of the text query $\phi_\text{quer}$, along with a set of canonical phrases $\phi_\text{canon}^i$. We compute cosine similarity between the rendered embedding and canonical phrase embeddings, then compute the pairwise softmax between the rendered embedding text prompts. The relevancy score is then
$\text{min}_i~\frac{\text{exp}(\phi_\text{lang}\cdot\phi_\text{quer})}{\text{exp}(\phi_\text{lang}\cdot\phi_\text{canon}^i)+\text{exp}(\phi_\text{lang}\cdot\phi_\text{quer}))}$.
Intuitively, this score represents how much closer the rendered embedding is towards the query embedding compared to the canonical embeddings. All renderings use the same canonical phrases: \textit{``object"}, \textit{``things"}, \textit{``stuff"}, and \textit{``texture"}. We chose these as qualitatively ``average" words for queries users might make, and found them to be surprisingly robust to queries ranging from incredibly specific to visual or abstract. We acknowledge that choosing these phrases is susceptible to prompt-engineering, and think fine-tuning them could be an interesting future work, perhaps incorporating feedback from user interaction about negative prompts they \textit{do not} consider relevant.

\textbf{Scale Selection}: For each query, we compute a scale $s$ to evaluate $F_\text{lang}$. To accomplish this, we generate relevancy maps across a range of scales 0 to 2 meters with 30 increments, and select the scale that yields the highest relevancy score. This scale is used for all pixels in the output relevancy map. We found this heuristic to be robust across a broad range of queries and is used for all the images and videos rendered in this paper.
This assumes relevant parts of a scene are the same scale, see Limitations Sec.~\ref{sec:limitations}.

\textbf{Visibility Filtering}: Regions of the scene that lack sufficient views, such as those in the background or near floaters, may generate noisy embeddings. To address this issue, during querying we discard samples that were observed by fewer than five training views (approximately 5\% of the views in our datasets).
\begin{figure}
    \centering
    \includegraphics[width=\linewidth]{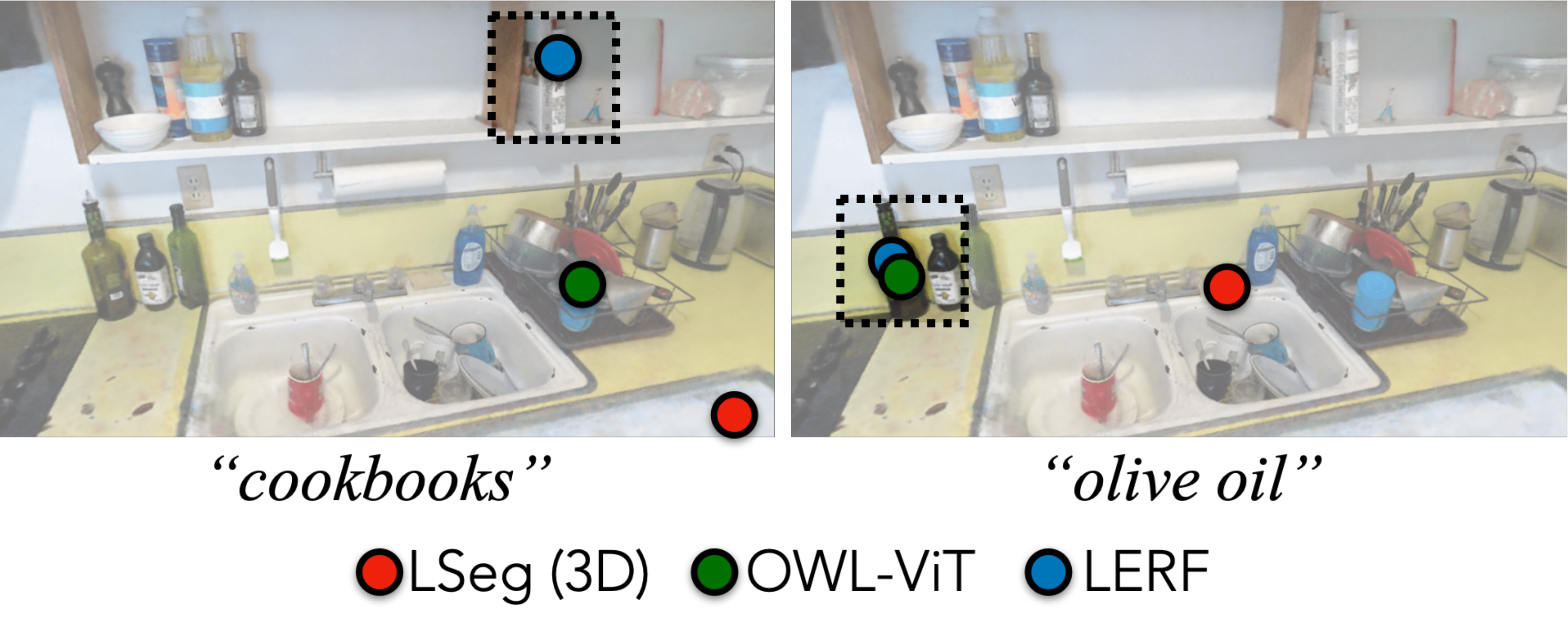}
    \caption{\textbf{Localization comparison} Qualitative comparison on localizing long-tail objects from novel views with LSeg in 3D (DFF) and OWL-ViT (Tab. \ref{tab:point_at})}
    \vspace{-15pt}
    \label{fig:localization_test}
\end{figure}
\subsection{Implementation Details}
We implement \algname{} in Nerfstudio~\cite{tancik2023nerfstudio}, on top of the Nerfacto method. Proposal sampling is the same except we reduce the number of \algname{} samples from 48 to 24 to increase training speed. We use the OpenClip~\cite{cherti2022reproducible} ViT-B/16 model trained on the LAION-2B dataset, with an image pyramid varying from $s_\text{min}=.05$ to $s_\text{min}=.5$ in 7 steps. The hashgrid used for representing language features is much larger than a typical RGB hashgrid: it has 32 layers from a resolution of 16 to 512, with a hash table size of $2^{21}$ and feature dimension of 8. The CLIP MLP used for $F_\text{lang}$ has 3 hidden layers with width 256 before the final 512 dimension CLIP output. The DINO MLP for $F_\text{DINO}$ has 1 hidden layer of dimension 256.

\looseness=-1 We use the Adam optimizer for proposal networks and fields with weight decay $10^{-9}$, with an exponential learning rate scheduler from $10^{-2}$ to $10^{-3}$ over the first 5000 training steps. All models are trained to 30,000 steps (45 minutes), although good results can be obtained in as few as 6,000(8 minutes) as presented in the Appendix. We train on an NVIDIA A100, which takes roughly 20GB of memory total. 
One can interactively query in real-time within the Nerfstudio viewer. The $\lambda$ used in weighting CLIP loss is $0.01$, chosen empirically and ablated in Sec \ref{sec:ablations}. When computing relevancy score, we multiply similarity by $10$ as a temperature parameter within the softmax.
\begin{figure}
    \centering
    \includegraphics[width=\linewidth]{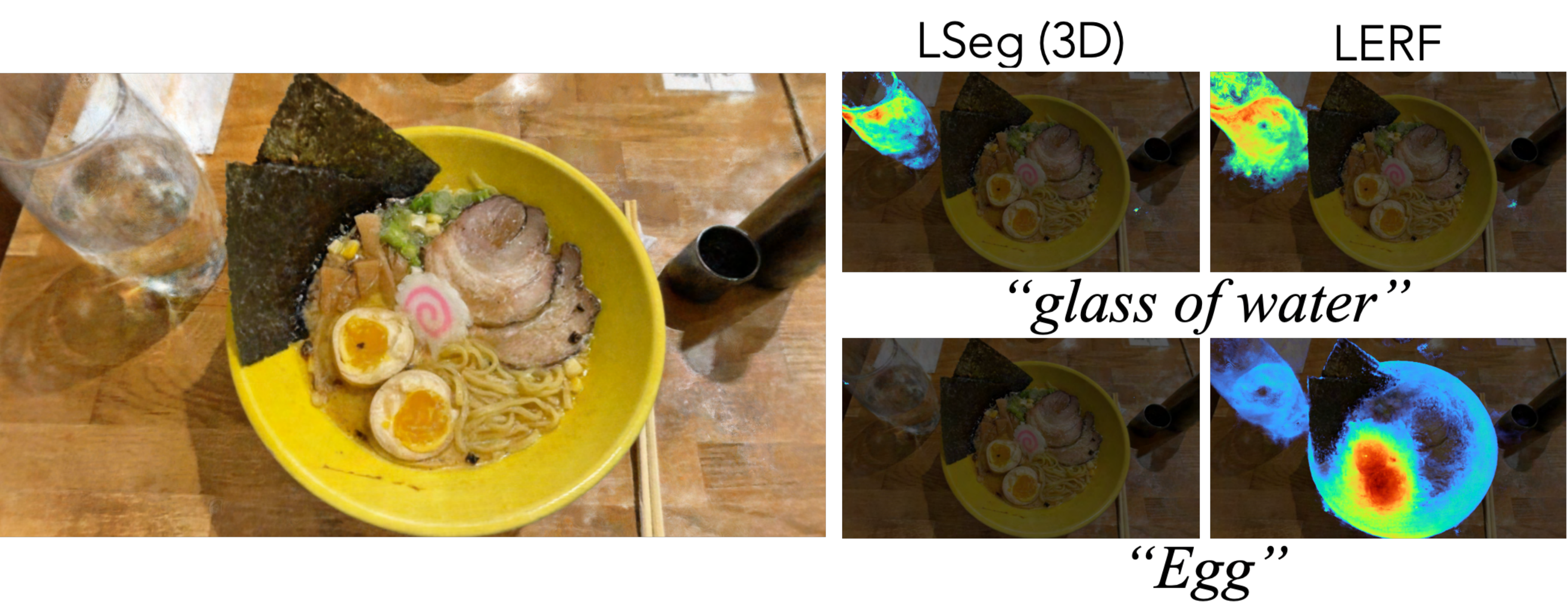}
    \caption{\textbf{Comparison to LSeg in 3D}: LSeg performs well on \textit{``glass of water"} since cups are in the COCO dataset, but cannot locate an out-of-distribution query like an egg.}
    \label{fig:lseg_compare}
\end{figure}

\section{Experiments}
We examine the capabilities of \algname{} and find that it can effectively process a wide variety of input text queries, encompassing various aspects of natural language specifications that current open-vocab detection frameworks encounter difficulty with. Though existing 3D scan datasets exist, they tend to be either of singulated objects~\cite{co3d,objaverse}, or are RGB-D scans without enough views to optimize high quality NeRFs~\cite{scannet}, and such simulated or scanned scenes contain few long-tail objects~\cite{straub2019replica}. Emphasizing the capability of \algname{} to handle real-world data, we collect 13 scenes containing a mixture of in-the-wild (grocery store, kitchen, bookstore) and posed long-tail (teatime, figurines, hand) scenes. We capture scenes using the iPhone app Polycam, which runs on-board SLAM to find camera poses, and use images of resolution 994$\times$738.
\begin{table}
\centering
\begin{tabular}{lccc}
\toprule
\textbf{Test Scene} & \textbf{LSeg (3D)
} & \textbf{OWL-ViT} & \textbf{\algname{}} \\
\midrule
waldo\_kitchen  & 13.0\% & 42.6\% & \textbf{81.5\%} \\
bouquet         & 50.0\% & 66.7\% & \textbf{91.7\%} \\
ramen           & 15.0\% & \textbf{92.5\%} & 62.5\% \\
teatime         & 28.1\% & 75.0\% & \textbf{93.8\%} \\
figurines       & 8.9\%  & 38.5\% & \textbf{79.5\%} \\
\midrule
Overall & 18.0\% & 54.8\% & \textbf{80.3\%} \\
\bottomrule
\end{tabular}
\caption{\textbf{Localization accuracy} comparison between Distilled Feature Fields using LSeg, OWL-ViT, and \algname{}. Overall performance is calculated by aggregating scene results. Refer to supplements for more details on scenes and text queries.}
\vspace{-15pt}
\label{tab:point_at}
\end{table}

\subsection{Qualitative Results}
\label{sec:qualitative}
We visualize relevancy score by normalizing the colormap for each query from 50\% (less relevant than canonical phrases) to the maximum relevancy. Extensive visualizations of all scenes can be found in the Appendix, and in Fig.~\ref{fig:results} we select 5 representative scenes which demonstrate \algname{}'s ability to handle natural language. Visual comparison with LSeg in 3D are presented in Fig~\ref{fig:lseg_compare}.

\algname{} captures language features of a scene at different levels of detail, supporting queries of properties like \textit{``yellow"}, as well as highly specific queries like names of books and specific characters from TV shows (\textit{``jake from adventure time"}).
Because of the lack of discrete categories, objects can be relevant to multiple queries: in the figurine scene, abstract text queries can create semantically meaningful groupings. \textit{``Cartoon"} selects the cat, Jake, rubber duck, miffy, waldo, toy elephant. \textit{``Bath toy"} selects rubber-like objects, such as rubber duck, Jake, and toy elephant (made of rubber). Toy elephant is strongly highlighted for three different queries, demonstrating the ability of \algname{} to support different semantic tags for the same object.

\begin{figure}
    \centering
    \includegraphics[width=\linewidth]{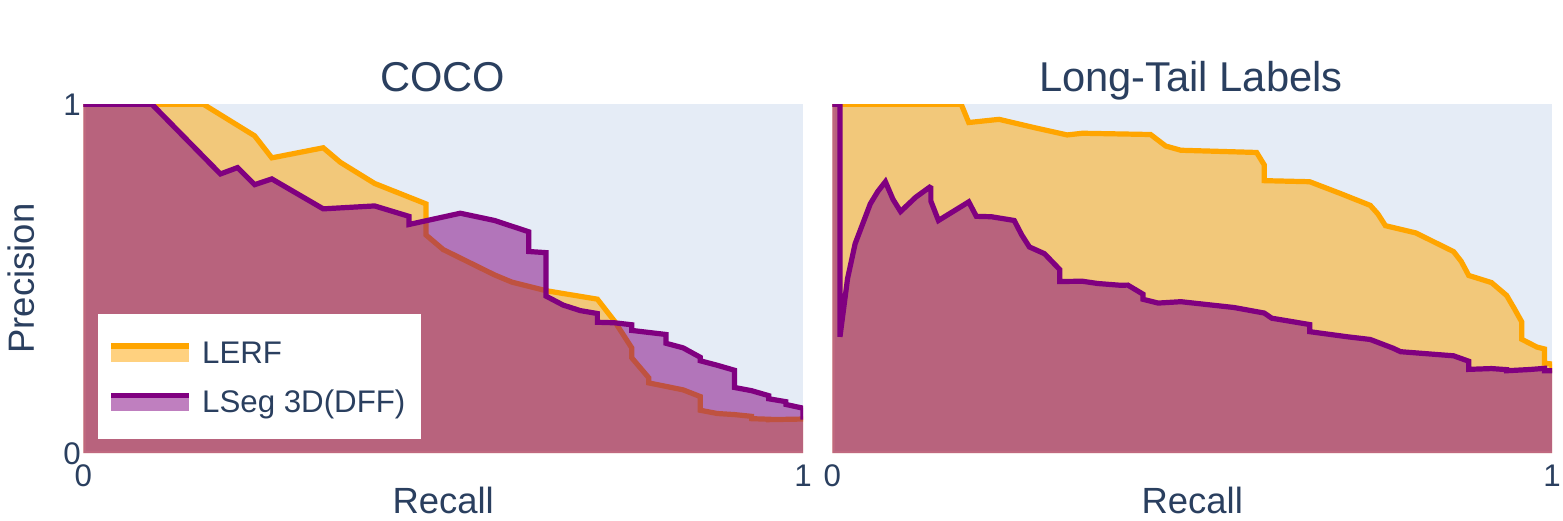}
    \caption{Precision recall curves for 3D existence experiments, Sec \ref{sec:existence}. LSeg performs similarly to \algname{} on in-distribution labels, but significantly suffers on long-tail labels of wild scenes.}
    \vspace{-10pt}
    \label{fig:existence_pr}
\end{figure}

\subsection{Existence Determination}
\label{sec:existence}
We evaluate if \algname{} can detect whether an object exists within a scene. We label ground truth existence for 5 scenes, collecting two sets of labels: 1) labels from COCO to represent in-distribution objects to LSeg and 2) our own long-tail labels, which consist of queries of 15-20 objects in each scene concatenated together, for 81 total queries.
See the Appendix for all queries. 
\algname{} determines whether an object exists in the scene by rendering a dense pointcloud over visible geometry, and returns ``True" if any point has a relevancy score over a threshold.

We compare against distilling LSeg features into 3D as in DFF~\cite{kobayashi2022decomposing}, but implemented in our own codebase for an apples-to-apples comparison. We remove scale as a parameter to $F_\text{lang}$ for LSeg since it outputs pixel-aligned features. We report precision-recall curves over relevancy score thresholds in Fig.~\ref{fig:existence_pr}. This experiment reveals that LSeg, trained on limited segmentation datasets, lacks the ability to represent natural language effectively. Instead, it only performs well on common objects that are within the distribution of its training set, as demonstrated in Fig.~\ref{fig:lseg_compare}.

\subsection{Localization}
To evaluate how well \algname{} can localize text prompts in a scene we render \textit{novel} views and label bounding boxes for 72 objects across 5 scenes. For 3D methods we consider a label a success if the highest relevancy pixel lands inside the box, or for OwL-ViT if the center of the predicted box does. Results are presented in Table~\ref{tab:point_at} and Fig.~\ref{fig:localization_test}, and suggest that language embeddings embedded in \algname{} strongly outperform LSeg in 3D for localizing relevant parts of a scene. We also compare against the 2D open-vocab detector OWL-ViT by rendering full-HD NeRF views and selecting the bounding box with the highest confidence score for the text query. OwL-ViT outperforms LSeg in 3D, but suffers compared to \algname{} on long-tail queries.

\subsection{Ablations}
\label{sec:ablations}
\textbf{No DINO}: Removing DINO results in a qualitative deterioration in the smoothness and boundaries of relevancy maps, especially in regions with few surrounding views or little geometric separation between foreground and background. We show two illustrative examples where DINO improves the quality of relevancy maps in Fig.~\ref{fig:ablations}.

\textbf{Single-Scale Training}: We ablate multi-scale CLIP supervision from the pipeline by only training on a fixed $s_0=15\%$ image scale. Doing so significantly impairs \algname{}'s ability to handle queries of \textit{all} scales, failing on both large (\textit{``espresso machine"}) queries it doesn't have enough context for, as well as queries for which it does (\textit{``creamer pods"}). These results imply that multi-scale training regularizes the language field at all scales, not just ones with relevant context for a given query.

\begin{figure}
    \centering
    \includegraphics[width=\linewidth]{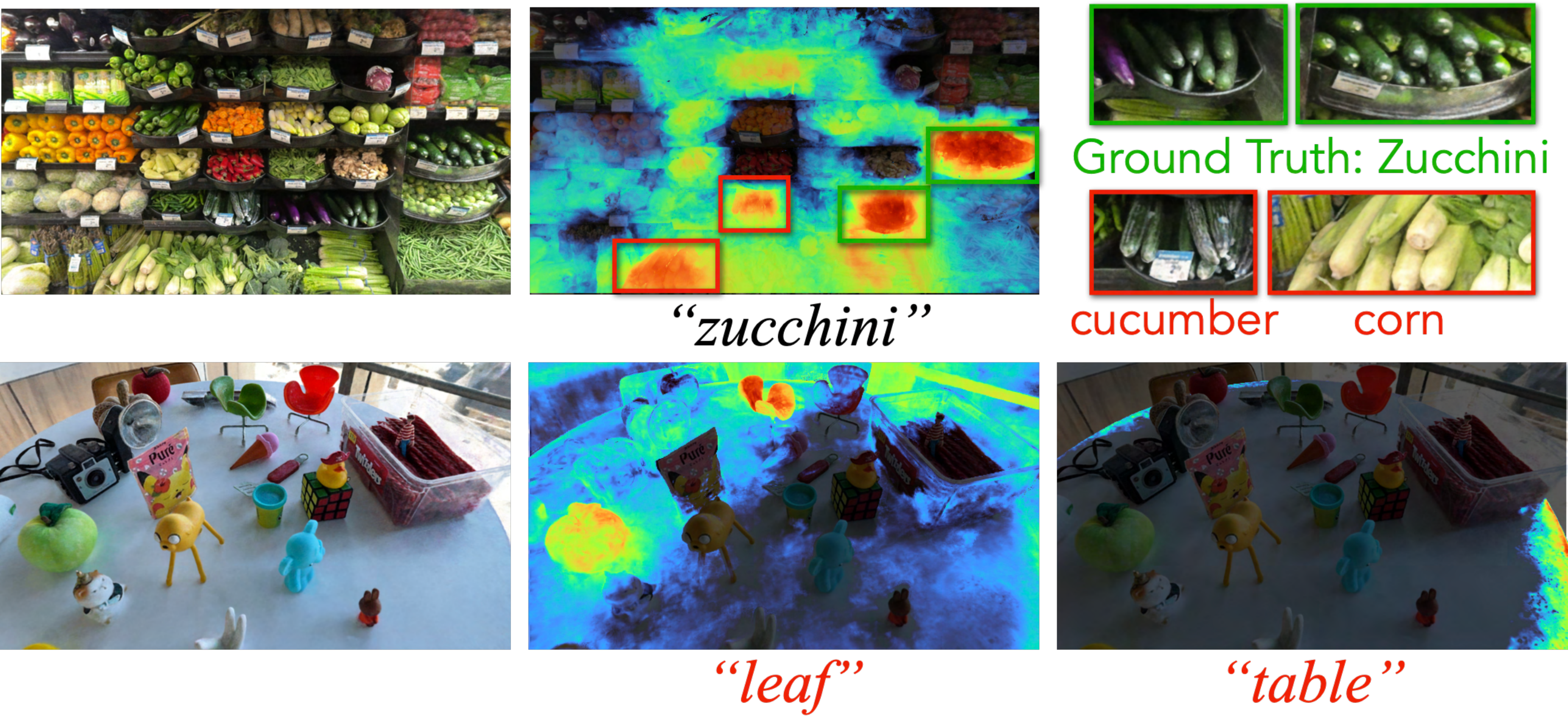}
    \caption{\textbf{Failure cases}: \algname{} struggles with identifying objects that appear visually similar to the query: \textit{``Zucchini"} also activates on other long, green-ish vegetables, and \textit{``leaf"} activates on the green plastic chair. \algname{} also struggles with global/spatial reasoning, where \textit{``table"} only activate on the edges of the table.}
    \vspace{-15pt}
    \label{fig:limitations}
\end{figure}
\section{Limitations}
\label{sec:limitations}
\algname{} has limitations associated with both CLIP and NeRF; some are visualized in Fig.~\ref{fig:limitations}. 
Like CLIP, language queries from \algname{} often exhibit ``bag-of-words" behavior (i.e., \textit{``not red"} is similar to \textit{``red"}) and struggles to capture spatial relationships between objects. \algname{} can be prone to false positives with queries that appear visually or semantically similar: \textit{``zucchinis"} activate on other similarly-shaped vegetables, though zucchinis are more relevant than the distractors (Fig.~\ref{fig:limitations}).

\algname{} requires known calibrated camera matrices and NeRF-quality multi-view captures, which aren't always available or easy to capture. The quality of language fields is bottlenecked by the quality of the NeRF recontsruction. In addition, because of the volumetric input to $F_\text{lang}$, objects which are near other surfaces without side views can result in their embeddings being blurred to their surroundings since no views see the background without the object. This results in similar blurry relevancy maps to single-view CLIP (Fig.~\ref{fig:clip-3d-comparison}). In addition, we only render language embeddings from a single scale for a given query. Some queries could benefit from or even require incorporating context from multiple scales (eg \textit{``table"}).

\section{Conclusions}
We present \algname{}, a novel method of fusing raw CLIP embeddings into a NeRF in a dense, multi-scale fashion without requiring region proposals or fine-tuning. We find that it can support a broad range of natural language queries across diverse real-world scenes, strongly outperforming pixel-aligned LSeg in supporting natural language queries. \algname{} is a general framework that supports any aligned multi-modal encoders, meaning it can naturally support improvements to vision-language models. Code and datasets will be released after the submission process.

\section{Acknowledgements}
This material is based upon work supported by the National Science Foundation Graduate Research
Fellowship Program under Grant No. DGE 2146752. Any opinions, findings, and conclusions
or recommendations expressed in this material are those of the author(s) and do not necessarily reflect the views of the National Science Foundation. The authors were supported in part by equipment grants from NVIDIA. We thank our colleagues who provided helpful feedback and suggestions, in particular Jessy Lin, Simeon Adebola, Satvik Sharma, Abhik Ahuja, Rohan Mathur, and Alexander Kristoffersen for their helpful feedback on paper drafts, and Boyi Li, Eric Wallace, and members of the Nerfstudio team for their generous conversations about the project.

{\small
\bibliographystyle{ieee_fullname}
\bibliography{egbib}
}

\clearpage
\appendix
\section{Videos}
We provide videos illustrating queries in 9 scenes on our project website. In all videos, the relevancy map images are post-processed such that all pixel values with relevancy score less than 0.5 are fully transparent. We choose this threshold because relevancy values under 0.5 means the rendered language embedding is more similar to the canonical negative phrases used (``object", ``things", ``stuff", ``texture") than to the positive query prompt, so we consider them irrelevant to the query. This relevancy map is overlaid on the RGB renders. Relevancy map scale selection and normalization factor is constant across the entire video sequence to show scene-wide 3D consistency. During post-process editing we select queries which are visible during the specific segment of video; but we also  provide full raw, uncut videos of outputs for the queries in the kitchen scene to give readers an idea of the scene-wide activations. Many other scenes contain wide-angle shots to observe relevancy maps scene-wide.

One notable property of relevancy maps across the scene that is more apparent in videos is that regions similar to, but not matching the query are also assigned a non-zero relevancy score, though not as high. For example, the query for \textit{``utensils"} highlights most on the utensils in the dish drainer, but also on the knives hanging on the wall and utensils in the sink. The query for \textit{``wooden spoon"} is most activated on the spoon, but also activates on other wooden components of the scene. This could be viewed as either a positive aspect of the language field in that it naturally groups similar regions to a query together, or a downside in that it can provide too many relevant regions in addition to the highest activation. For example, for the \textit{``refrigerator"} query, the highest activation is assigned to the refrigerator, but much of the remaining kitchen space is also labeled as relevant. We hypothesize that this is related to the lack of grounding with visual-language models like CLIP. 
We find that the longer-tail and more specific the query, the more separation it tends to have with canonical phrases and hence the result more obviously pops out from other objects. Another effect visible in videos is the presence of ``floaters" in the scene which can produce spurious activations, as discussed at length in Fig.\ref{fig:poor_geometry} and Sec.\ref{sec:more_limitations}.

\begin{figure}
    \centering
    \includegraphics[width=\linewidth]{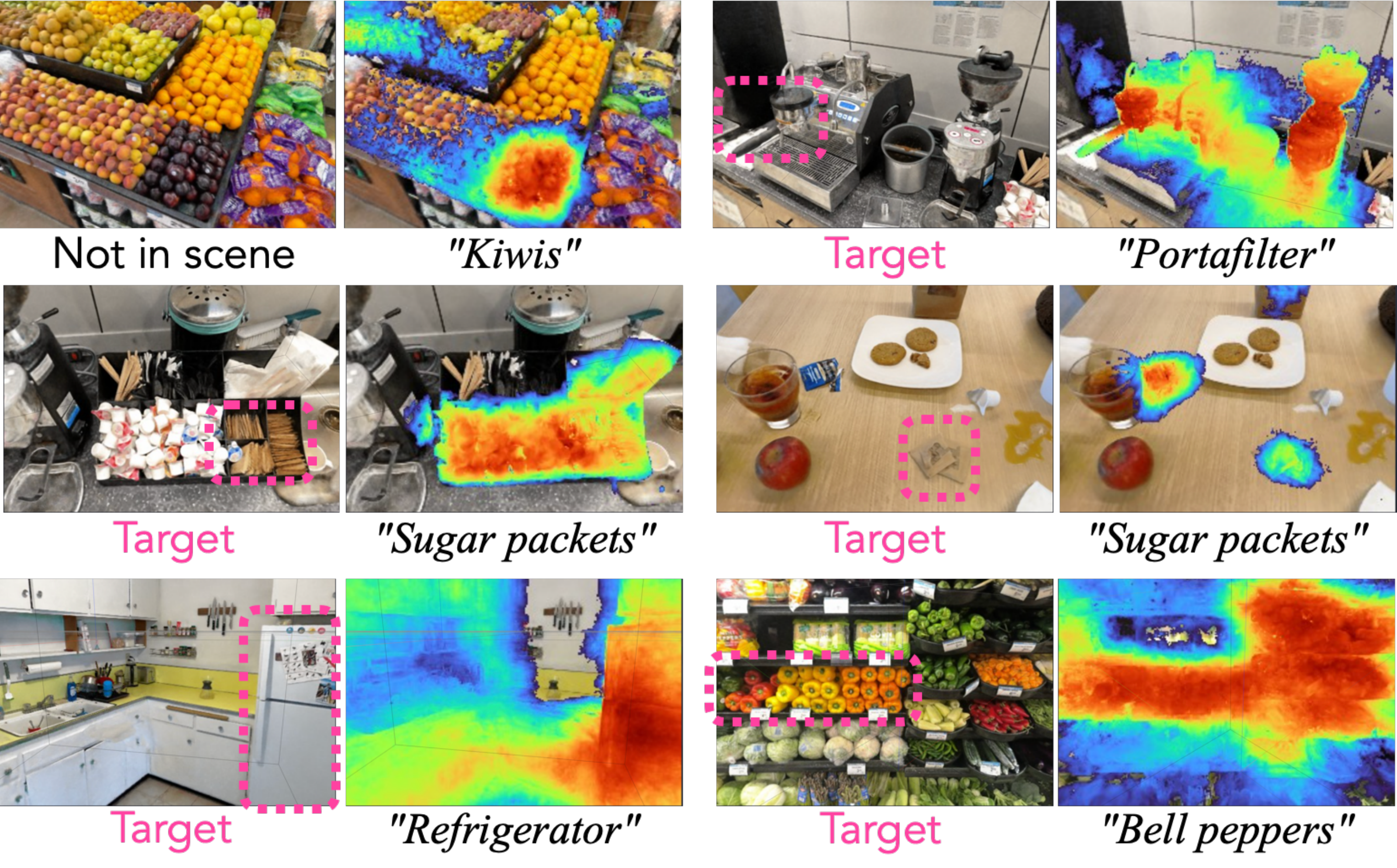}
    \caption{\textbf{Language and visual ambiguities from CLIP}: Cases with incorrect relevancy renders. Some failures can be attributed to visual similarity to the query (eg \textit{``bell peppers"} gets distracted on jalepenos, \textit{``portafilter"} activates on the grinder spout which has a similar metal cylindrical appearance, and \textit{``refrigerator"} slightly activates on the white rectangular cabinets). Others are more flat-out failures, with \textit{``sugar packets"} seemingly a confusing case to detect, and \textit{``kiwis"} activating strongly on plums rather than correctly predicting nothing.}
    \label{fig:ambiguity}
\end{figure}

\section{Additional qualitative results}
We present a more complete list of results from scenes not pictured in the main text or videos in Fig.~\ref{fig:more_results}. 

\begin{figure*}
    \centering
    \includegraphics[width=\linewidth]{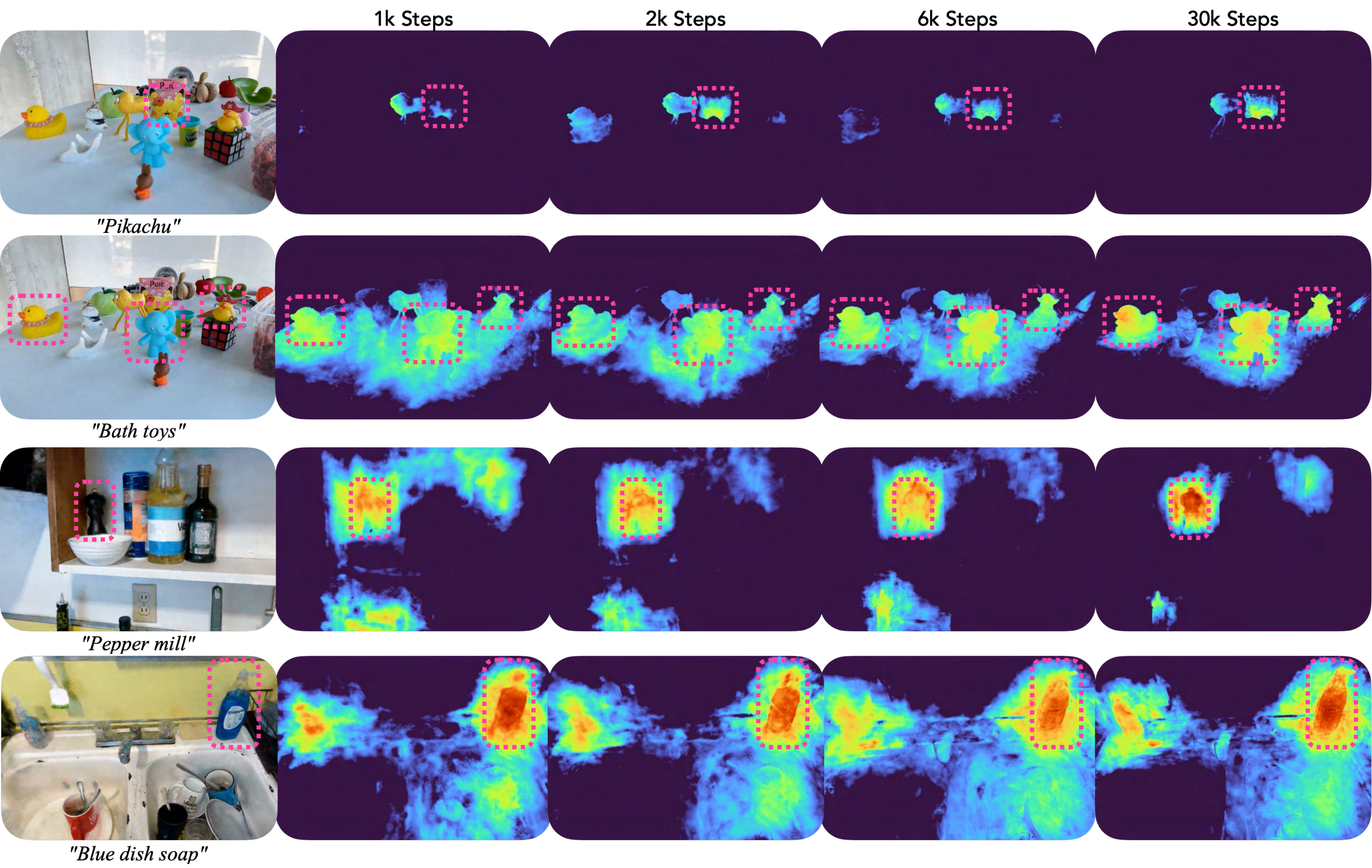}
    \caption{\textbf{\algname{} Convergence}. We visualize rendered relevancy maps at 1k, 2k, 6k, and 30k optimization steps. Relatively speaking, regions with more common semantics (like \textit{``blue dish soap"}) and more expansive multi-view converge faster, while more fine-grained properties (like \textit{``bath toys"}) take more steps. Notably, the optimization is quite stable: all queries produce reasonable activations within the first few thousand steps, and relevancy continues to refine over time.}
    \label{fig:speed}
\end{figure*}
\section{Numerical relevancy scores}
We explore the reliability of using relevancy as a threshold for existence determination in and report precision-recall curves. Here, we provide raw relevancy scores for the queries in Fig. 1 of the main text to illustrate the behavior across different types of queries. Scores are shown in Table~\ref{tab:splash_relevancy}. One can observe that highly descriptive queries including visual and semantic properties produce higher relevancy scores (eg \textit{``blue dish soap"}), and the lowest are abstract queries like \textit{``electricity"} or small objects in clutter with not many close-up views (\textit{``wooden spoon"}).

\section{Convergence speed}
Videos and images for \algname{} were rendered after 30k optimization steps. However, usable relevancy maps can be obtained much sooner, as this section explores. We visualize relevancy maps and RGB renders of the kitchen scene and figurines scene after 1k, 2k, 6k, and 30k steps in Fig. \ref{fig:speed}. Because density converges significantly in NeRF within the first thousand steps, language embeddings in 3D are already usable at this stage. However, some fine-grained queries or small objects suffer in performance until later in optimization. In the \textit{``Pikachu"} query, early training steps confuse the yellow figurine (Jake from Adventure Time) which is visually similar. As \algname{} converges, the actual Pikachu in the scene has higher relevancy than Jake. For fine-grained properties like \textit{``bath toys"}, the relevancy starts out more blurry and becomes sharper and more isolated to the correct objects over time. Objects without much geometric separation like \textit{``pepper mill"} also take longer to converge, since the surrounding geometry can be less precise.
\begin{table}[]
    \centering
    \begin{tabular}{cc}
       \toprule
       Text query & Maximum relevancy score \\
       \midrule
       utensils  &  0.77\\
       wooden spoon  &  0.60\\
       blue dish soap  &  0.83\\
       waldo  &  0.76\\
       paper towel roll  &  0.75 \\
       electricity  &  0.63\\
       yellow  &  0.73\\
       boops  &  0.77\\
       \bottomrule
    \end{tabular}
    \caption{Maximum relevancy scores for each text query in Fig. 1 of main text, calculated from the displayed viewpoint. Highly specific queries have a higher relevancy value (\textit{``blue dish soap"}, 0.83), while abstract queries can have lower ones (\textit{``electricity"}, 0.63).
    }
    \label{tab:splash_relevancy}
\end{table}

\section{Experiment details}
We provide a list of the labels used for the localization experiment in Tab. \ref{tab:box_labels}. Each label was labeled in 3-4 different views. We provide an exhaustive list of our custom long-tail labels for the existence experiment in Tab. \ref{tab:existence_labels}. 

\begin{figure}
    \centering
    \includegraphics[width=.95\linewidth]{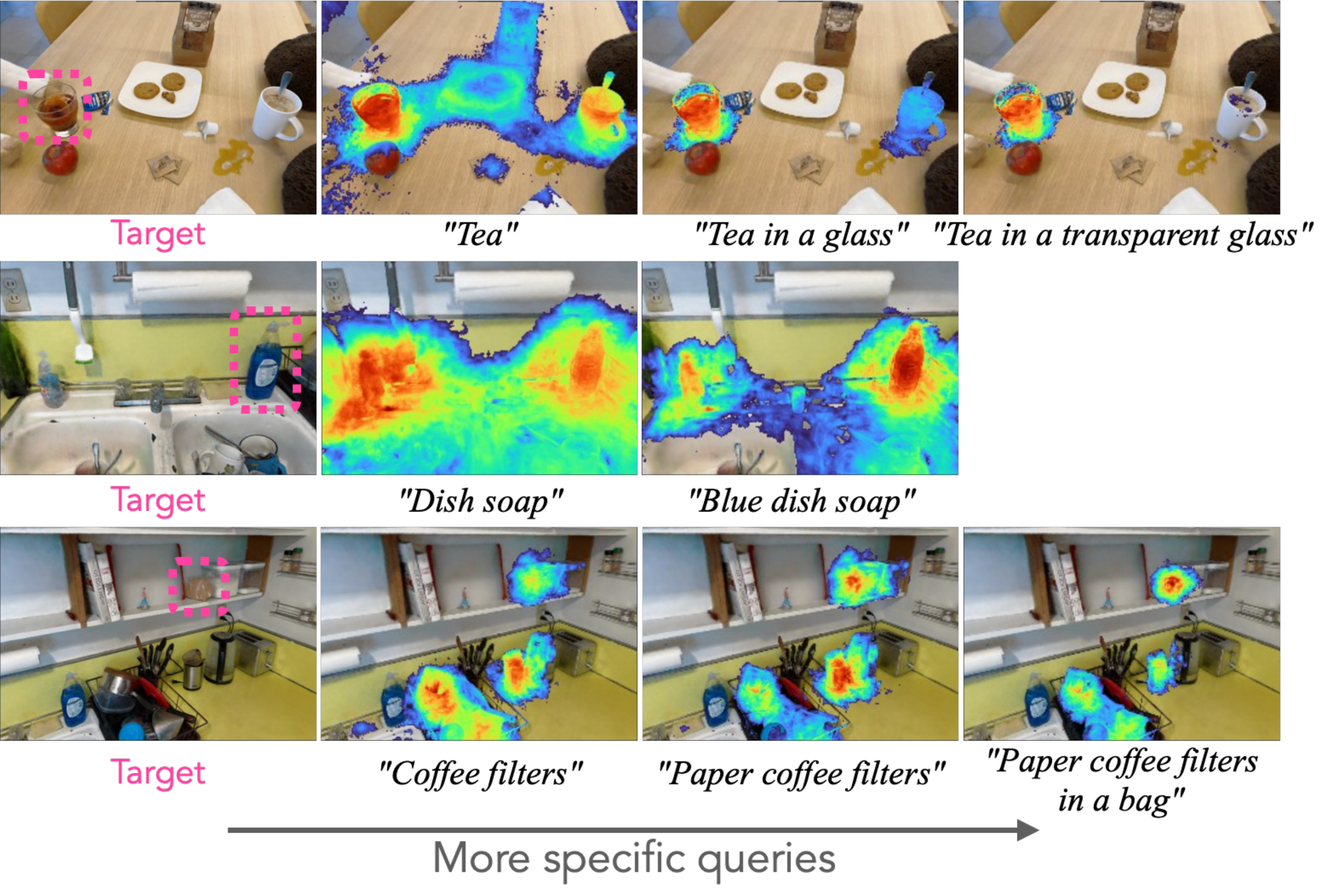}
    \caption{\textbf{Prompt tuning case study}: Some objects are sensitive to the prompt, with more specific wordings producing better results.}
    \label{fig:prompt_tuning}
\end{figure}

\begin{figure}
    \centering
    \includegraphics[width=.95\linewidth]{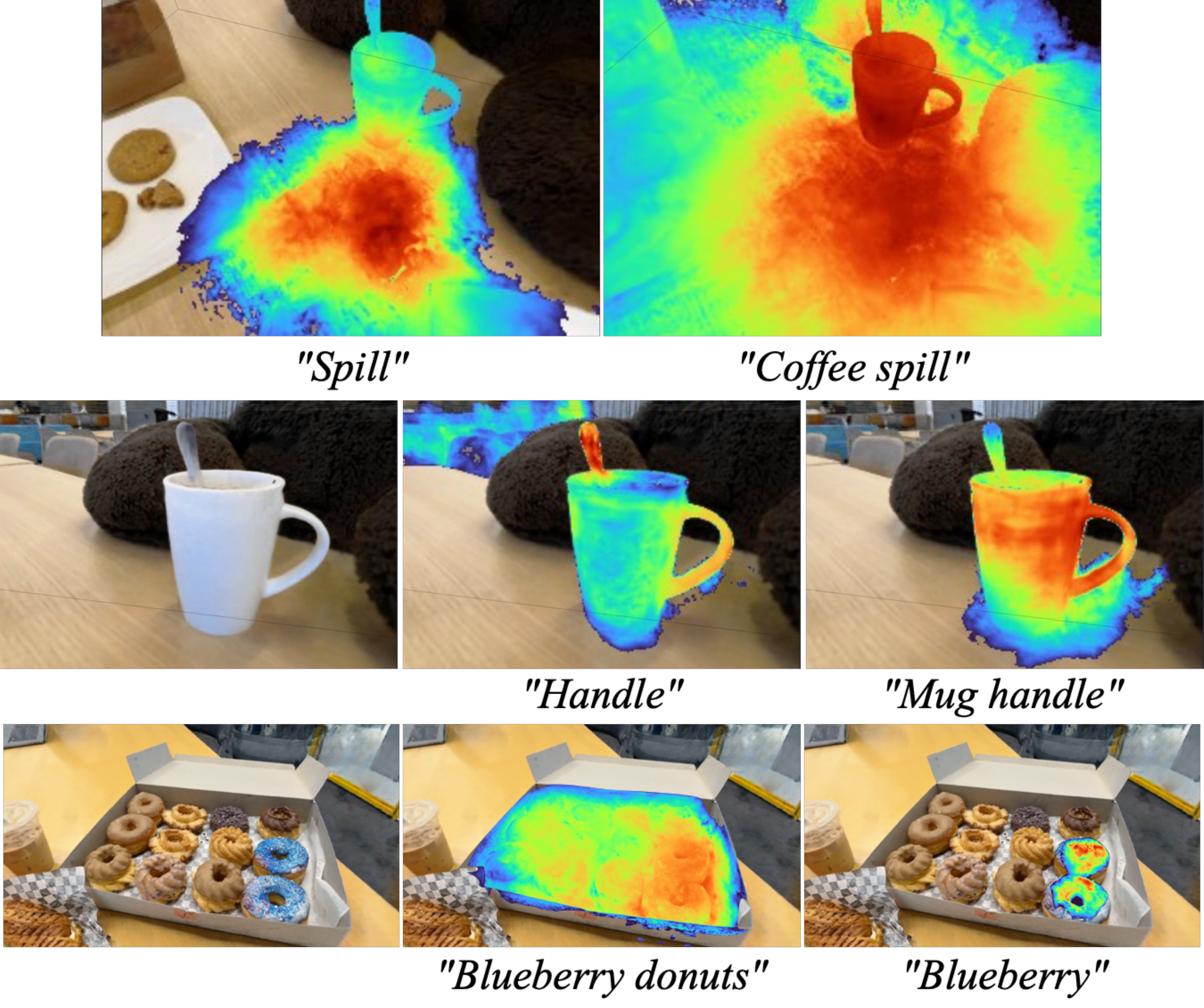}
    \caption{\textbf{CLIP bag-of-words behavior}: CLIP sometimes behaves as a bag-of-words, resulting in some adjectives not properly incorporating into queries.}
    \label{fig:bagofwords}
\end{figure}

\begin{figure}
    \centering
    \includegraphics[width=.95\linewidth]{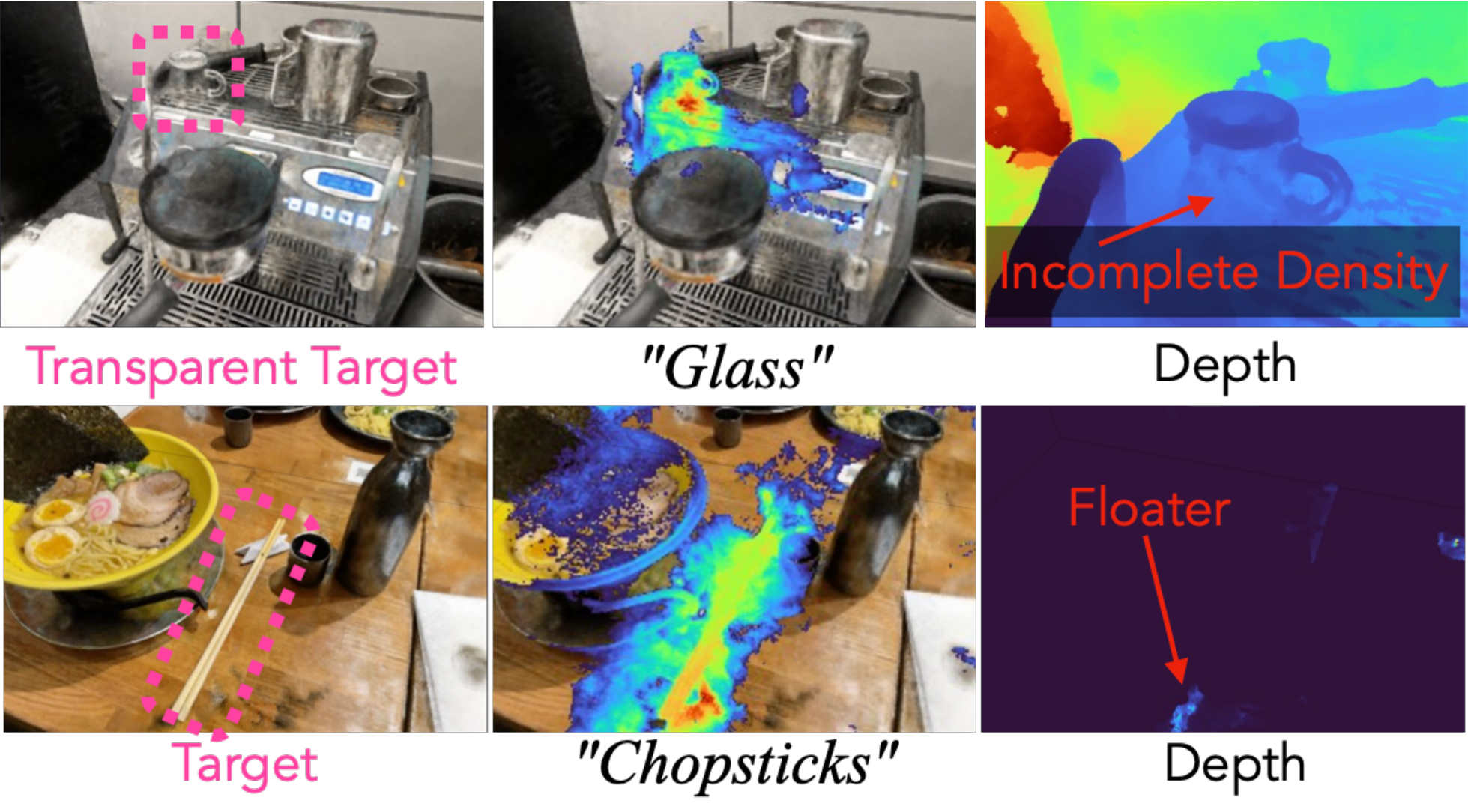}
    \caption{\textbf{Degradation with poor NeRF geometry}: Floaters and incomplete geometry can produce unreliable rendered CLIP embeddings.}
    \label{fig:poor_geometry}
\end{figure}

\begin{figure}
    \centering
    \includegraphics[width=.95\linewidth]{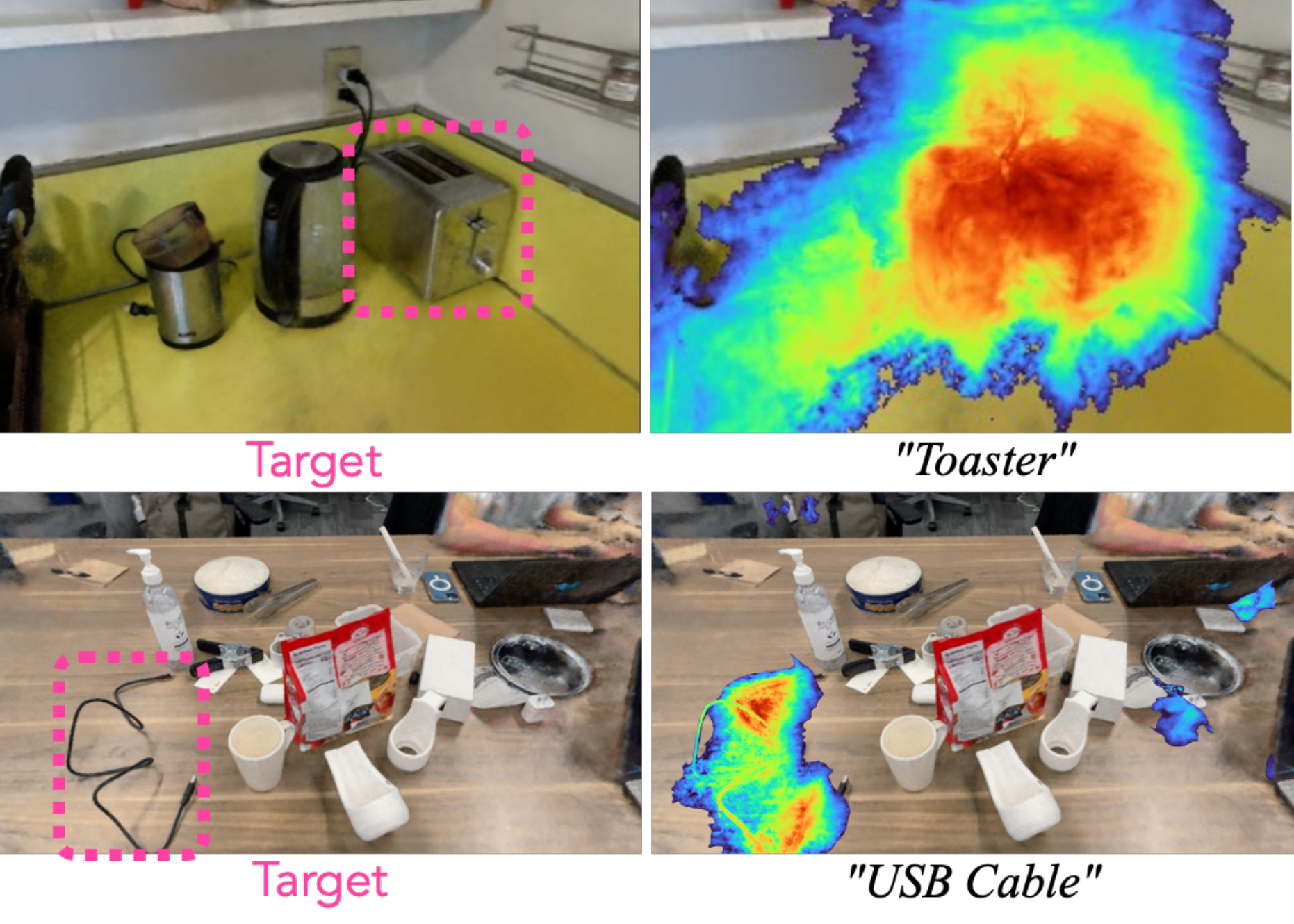}
    \caption{\textbf{Geometric separation impacts quality}: Queries without much geometric separation can blur between objects and foreground-background. In the toaster case, very few viewing angles were taken because of its position, which results in a fuzzier boundary.}
    \label{fig:geometry_blurring}
\end{figure}
\section{Detailed Illustrations of Limitations}
\label{sec:more_limitations}
\algname{} inherits limitations from CLIP relating to language ambiguity and prompt sensitivity, as well as from NeRF's geometry representation capabilities. We present additional figures on failure cases to complement the ones provided in the main text.

Fig.\ref{fig:ambiguity} showcases visual and language ambiguity from our usage of CLIP. Some queries get confused by unrelated regions of the scene because they appear very similar, such as the portafilter and the coffee grinder. In the refrigerator query, unrelated parts of the kitchen also activate in relevancy maps, though less strongly than refrigerator, because the CLIP embeddings of square white cabinets are more similar to a refrigerator than the canonical phrases. Sugar packets appear to be a confusing case for \algname{}, getting distracted in two separate scenes (teatime, espresso machine) with a tea packet and creamer pods respectively.

Fig.\ref{fig:bagofwords} highlights another well-known undesirable property inherited from CLIP: text embeddings often behave as a bag-of-words rather than a grammatically parsed sentence. As a result, sometimes adding additional adjectives cause the output to latch onto incorrect regions (\textit{``mug handle"} vs \textit{``handle"} or \textit{``coffee spill"} vs \textit{``spill"}.)

Fig.\ref{fig:poor_geometry} shows performance degradation when geometry is unreliable in the underlying NeRF: reflective objects like the table in the ramen scene can produce holes, which result in CLIP embeddings from multiple views incorrectly averaging. Highly transparent objects like the glass cup in the espresso scene also suffer from lack of density, since the rendering weights mostly focus on the opaque background rather than the transparent foreground. 

Fig.\ref{fig:prompt_tuning} illustrates examples of relevancy maps improving in quality with subtly changed queries to become more and more specific. Usually this has a subtle effect on relevancy maps by refining the activation to a more localized region, for example providing progressively more descriptive queries improves the relevancy activation on the tea cup. This effect can also be drastic, for example \textit{``Dish soap"} primarily activates on a pump soap bottle, but describing \textit{``blue dish soap"} shifts focus to the correct object.

Finally, Fig.\ref{fig:geometry_blurring} shows cases where lack of geometric separation (a cable close to a table, or a toaster flush in the corner) causes the relevancy maps to blur into other surrounding objects because most views of the background contain the foreground object in front.






\begin{table*}[t!]
    \centering
    \begin{tabular}{lcccc}
    Scene & \multicolumn{4}{c}{Text queries} \\
    \toprule
                & blue hydroflask & coffee grinder & cookbooks & cooking tongs \\
                & copper-bottom pot & dish soap & faucet & knives \\
Kitchen  & olive oil & paper towel roll & pepper mill & pour-over vessel \\
                & power outlet & red mug & scrub brush & sink \\
                & spice rack & utensils & vegetable oil & waldo \\
\midrule
                & big white crinkly flower & bouquet & carnation & daisy \\
Bouquet         & eucalyptus & lily & rosemary & small white flowers \\
                & vase & & & \\
\midrule
                & green apple & ice cream cone & jake & miffy \\
                & old camera & pikachu & pink ice cream & porcelain hand \\
Figurines       & quilted pumpkin & rabbit & red apple & rubber duck \\
                & rubics cube & spatula & tesla door handle & toy cat statue \\
                & toy chair & toy elephant & twizzlers & waldo \\
\midrule
                & bowl & broth & chopsticks & egg \\
Ramen           & glass of water & green onion & napkin & nori \\
                & pork belly & ramen & sake cup & wavy noodles \\
\midrule
                & bag of cookies & bear nose & coffee & coffee mug \\
Teatime         & cookies on a plate & dall-e & hooves & paper napkin \\
                & plate & sheep & spill & spoon handle \\
                & stuffed bear & tea in a glass & yellow pouf & \\
                \bottomrule
    \end{tabular}
    \caption{Labels used during detection experiments (75 total).}
    \label{tab:box_labels}
\end{table*}

%
\begin{table*}[]
    \centering
    \begin{tabular}{cc}
        \toprule
         Scene & Positive Labels \\
         \midrule
         Figurines & jake, miffy, rabbit, bunny, old camera, toy elephant, twizzlers, quilted pumpkin, tesla door handle, \\
         & porcelain hand, rubics cube, rubber duck, apple, ice cream cone, \\ & pink ice cream, toy cat statue, toy chair, waldo, spatula, pikachu,table,\\
         \midrule
         Kitchen & red mug, pour-over vessel, olive oil, vegetable oil, cookbooks, waldo, \\ & dish soap, plates, sink, faucet, copper-bottom pot, utensils, knives, spice rack, \\
         & coffee grinder,flour, blue hydroflask, pepper mill, paper towel roll, scrub brush, \\ & power outlet, cooking tongs, transparent tupperware, coffee mug, coffee, spoon handle, \\
         \midrule
         Teatime & stuffed bear, sheep, bear nose, coffee mug, spill, tea in a glass, cookies on a plate, bag of cookies, dall-e, \\
         & hooves,coffee, yellow pouf, spoon handle, paper napkin, plate, wood, bag of food, hand sanitizer, mug,\\
         \midrule
         Table & wood texture, red bag of food, hand sanitizer bottle, airpods case, \\ & usb cable, brown paper napkins, transparent tupperware, colorful coaster, \\
         & packing tape roll, hardware clamps, \\ & iphone, laptop, metal cooking tongs, mug, drinking straw, table, tea in a glass,\\
         \midrule
         Bouquet & big white crinkly flower, bouquet, carnation, daisy, eucalyptus, lily, rosemary \\
         & small white flowers, table, flowers, pink flowers, wood, sink\\
         \bottomrule
    \end{tabular}
    \caption{Long-tail labels used for existence experiment. Each row shows the positive labels for a given scene. Negative labels consist of the positives for other scenes, re-labeled to match the ground truth.}
    \label{tab:existence_labels}
\end{table*}
\begin{figure*}
    \centering
    \includegraphics[width=.95\linewidth]{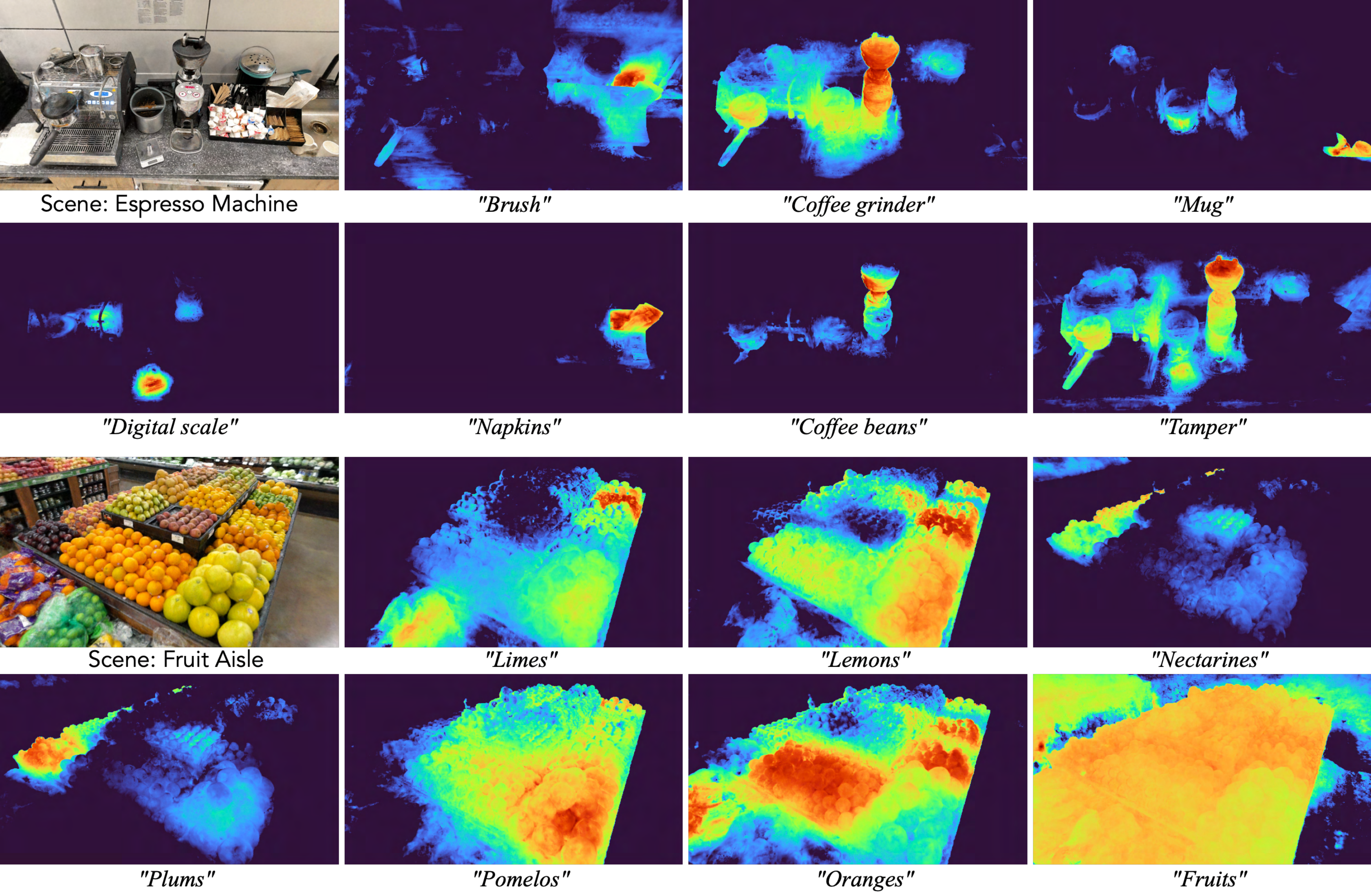}
    \includegraphics[width=.95\linewidth]{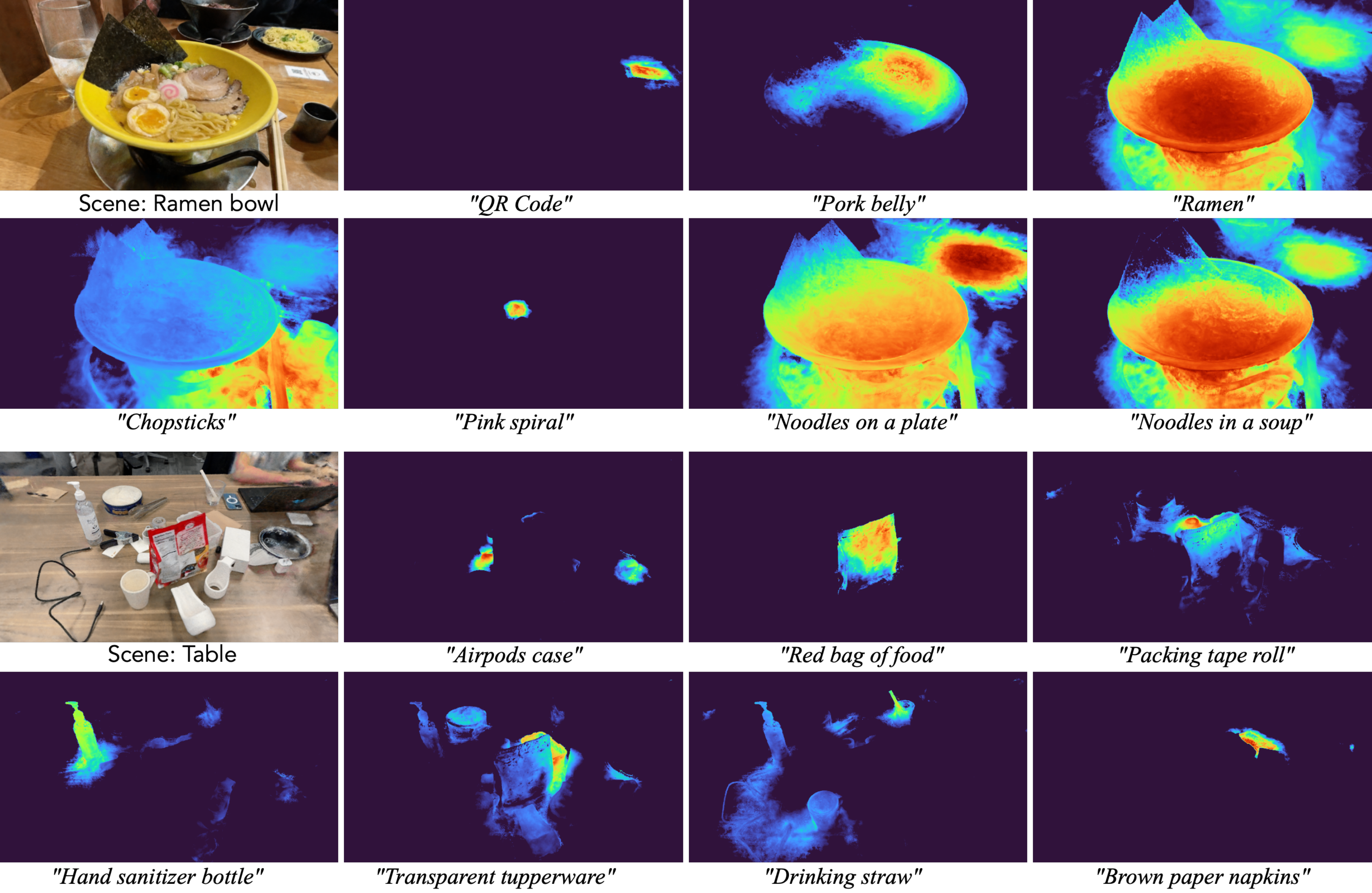}
    \caption{Additional results of scenes not reported fully in the main text or rendered in videos.}
    \label{fig:more_results}
\end{figure*}

\end{document}